\pdfoutput=1

\documentclass[11pt]{article}

\usepackage{EMNLP2022}

\usepackage{times}
\usepackage{latexsym}

\usepackage[T1]{fontenc}

\usepackage[utf8]{inputenc}

\usepackage{microtype}

\usepackage{inconsolata}

\usepackage{graphicx}
\usepackage{amsmath}
\usepackage{amssymb}
\usepackage{mathtools}
\usepackage{amsthm}
\usepackage{booktabs}

\usepackage{amsmath,amsfonts,bm}

\def\eqref#1{equation~\ref{#1}}

\def\1{\bm{1}}

\def\vb{{\bm{b}}}
\def\vc{{\bm{c}}}

\def\vx{{\bm{x}}}

\def\mC{{\bm{C}}}

\def\mH{{\bm{H}}}

\def\mK{{\bm{K}}}

\def\mQ{{\bm{Q}}}

\def\mV{{\bm{V}}}
\def\mW{{\bm{W}}}
\def\mX{{\bm{X}}}

\def\mZ{{\bm{Z}}}

\DeclareMathAlphabet{\mathsfit}{\encodingdefault}{\sfdefault}{m}{sl}
\SetMathAlphabet{\mathsfit}{bold}{\encodingdefault}{\sfdefault}{bx}{n}

\newcommand{\softmax}{\mathrm{softmax}}

\title{Simple Recurrence Improves Masked Language Models}

\author{Tao Lei \and Ran Tian \and Jasmijn Bastings \and Ankur P. Parikh\\
  Google Research \\
  \texttt{\{taole, tianran, bastings, aparikh\}@google.com} \\}

\begin{document}
\maketitle
\begin{abstract}
In this work, we explore whether modeling recurrence into the Transformer architecture can both be beneficial and efficient,
by building an extremely simple recurrent module into the Transformer.
We compare our model to baselines following the training and evaluation recipe of BERT.
Our results confirm that recurrence can indeed improve Transformer models by a consistent 
margin, without requiring low-level performance optimizations, and while keeping the number of parameters constant.
For example, our base model achieves an absolute improvement of 2.1 points averaged across 10 tasks and also demonstrates increased stability in fine-tuning over a range of learning rates.
\end{abstract}

\section{Introduction}

While the 
Transformer~\cite{transformer} relies solely on attention mechanisms 
for sequence modeling, many recent works have incorporated recurrence into the architecture and demonstrated superior performance in various applications.
For example, such modifications were shown to be beneficial for modeling long-range inputs~\cite{hutchins2022blockrecurrent}, accelerating language model training~\cite{lei-2021-attention} and improving translation and speech recognition systems~\cite{hao-etal-2019-modeling,srupp_speech,Chen2018TheBO}.

Even though combining attention and recurrence is useful in many cases, very little efforts have gone into language model pre-training and fine-tuning.
In particular, one open question is whether a combined model can be pre-trained 
and fine-tuned to achieve stronger accuracy compared to its attention-only counterparts.

\begin{figure}[t]
    \centering
    \includegraphics[width=2.6in]{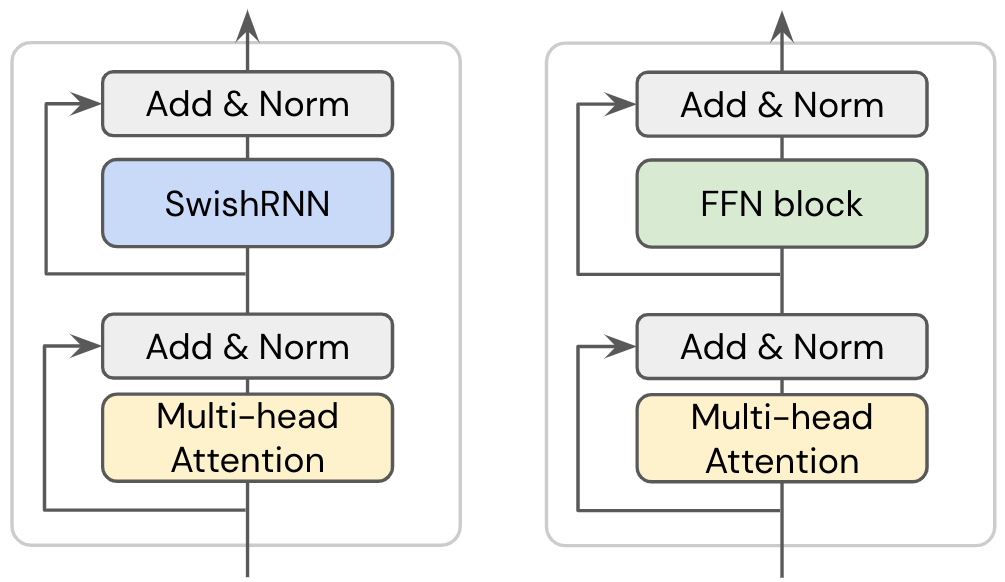}
    \caption{Our model architecture (left) and the original Transformer (right) for masked language modeling. We replace the feed-forward blocks with light-weight recurrence, which is interleaved with attention blocks.}
    \label{fig:architecture}
\end{figure}

We study this question in the case of masked language model training, specifically BERT~\cite{devlin-etal-2019-bert}.
Unlike previous work~\cite{huang2020transblstm}, we are interested in retaining the training efficiency of the model when combining attention and recurrence.
That is, the amount of parameters and computation should remain comparable to the baseline Transformer model.
However, making a recurrent model operating at a similar computation throughput as attention can be challenging, such as requiring CUDA implementations for GPUs~\cite{appleyard2016optimizing,bradbury2016quasi}.
To mitigate this issue, we propose a simple recurrent implementation which we call SwishRNN.
SwishRNN uses minimal operations in the recurrence step in order to accelerate computation, and can run on both TPUs and GPUs using a few lines of code in machine learning libraries such as Tensorflow.
We incorporate SwishRNN into BERT by 
substituting the feed-forward layers and keeping the same number of model parameters.

We pre-train our model and BERT baselines using the standard Wikipedia+Book corpus, and compare their fine-tuning performance on 10 tasks selected in the GLUE and SuperGLUE benchmark.
Our results confirm that modeling recurrence jointly with attention is indeed helpful, resulting in an average improvement of 2.1 points for the BERT-base models and 0.6 points for the large models.
The combined model also exhibits better stability, achieving more consistent fine-tuning results over a range of learning rates.

\section{Model}
In this section, we first give a quick overview of our model architecture and then describe the recurrence module SwishRNN in more details.

\subsection{Notation and Background}

The Transformer architecture interleaves a multi-headed attention block, $F_\textrm{att}$, with feed forward block, $F_\textrm{ffn}$, as shown in Figure~\ref{fig:architecture}. Between each block is a residual connection and layer normalization that we denote as ${F}_{\textrm{Add+Norm}}$. These functions are defined in the Appendix for completeness.

At each layer $k$, the hidden state of a Transformer is represented by an $l\times d$ matrix $\mX^k$, where $l$ is the sequence length and $d$ the hidden size\footnote{For simplicity of notation, the $l$ superscript is only included when necessary}. We define the intermediate hidden state $\bar{\mX}^k$ and input to the next layer $\mX^{k+1}$ as:
\begin{align}
\bar{\mX}^k &:= F_\textrm{Add+Norm} \left ( F_\textrm{att}(\mX^k), \mX^k \right ) \notag \\
\mX^{k+1} &:= F_\textrm{Add+Norm} \left ( F_\textrm{ffn}(\bar{\mX}^k), \bar{\mX}^k \right )
\label{eq:orig-transformer}
\end{align}

\subsection{Architecture}

Compared to the original architecture, we simply replace every feed-forward block $F_\textrm{ffn}$ with a recurrence block as shown in Figure~\ref{fig:architecture}.

\paragraph{SwishRNN}

\begin{figure}[t]
    \centering
    \includegraphics[width=2.4in,trim={1cm 0 0 0},clip]{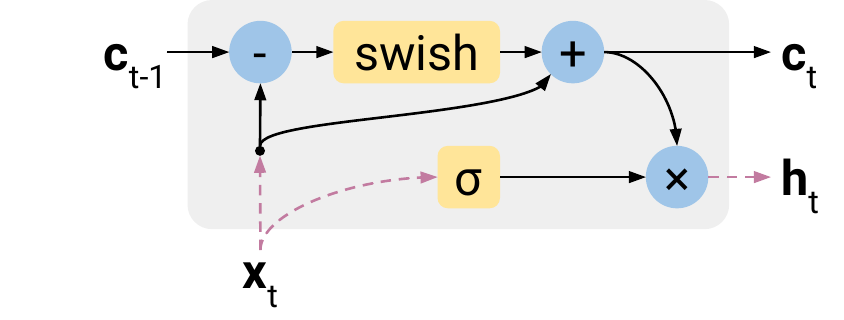}
    \vspace{-0.1in}
    \caption{The SwishRNN cell. A red dotted line represents a linear transformation.}
    \label{fig:cell}
\end{figure}
Modern accelerator hardwares such as TPUs and GPUs are highly optimized for matrix multiplications,
making feed-forward architectures such as attention very efficient.
Recurrent networks (RNNs) however involves sequential operations that cannot run in parallel.
In order to achieve a training efficiency comparable to the original Transformer, we use minimal sequential operations and demonstrate they are sufficient to improve the modeling power.

Specifically, SwishRNN uses only two matrix multiplications and an extremely simple sequential pooling operation. Let $\bar{\vx}[i] := \bar{\mX}[i,:]$ be the intermediate hidden vector of the $i$-th position (from Eq.~\ref{eq:orig-transformer}). SwishRNN first computes two linear transformations of $\bar{\mX}$:
\begin{align}
    \bar{\mX}_1 = \bar{\mX} \mW_1, \quad \bar{\mX}_2 = \bar{\mX} \mW_2
\end{align}
where $\mW_1$ and $\mW_2$ are $d\times d'$ parameter matrices optimized during training, $d$ is the input and output dimension of the model, and $d'$ is the intermediate dimension for recurrence.
The hidden vectors $\{\vc[i]\}_{i=1}^l$ are calculated as follows
\begin{align}
    \vc[i] = \texttt{Swish}\left(\vc[i\text{-}1] - \bar{\vx}_1[i]\right) + \bar{\vx}_1[i]
\end{align}
where $\texttt{Swish}()$ is the element-wise Swish activation function~\cite{ramachandran2018searching}.\footnote{$\texttt{Swish}(\vx) = \texttt{sigmoid}(\alpha\cdot \vx + \beta) \cdot \vx$. We initialize $\mathbf{\alpha}=\mathbf{1}$ and $\mathbf{\beta}=\mathbf{0}$ and optimize both scalar vectors during training.}
We use a $l\times d'$ matrix $\mC$ to represent the concatenated version of $\{\vc[i]\}_{i=1}^l$, and set $\vc[0]$ as an all-zero vector for simplicity.
Intuitively, step~(2) can be interpreted as a pooling operator where the greater value between $\vc[i\text{-}1]$ and $\bar{\vx}_1[i]$ are selected.\footnote{Note $\vc[i]=\bar{\vx}[i]$ if $\bar{\vx}[i] \gg \vc[i\text{-}1]$, and $\vc[i]=\mathbf{c}[i\text{-}1]$ if $\bar{\vx}[i] \ll \vc[i\text{-}1]$.}

The output vectors are obtained using a multiplicative gating similar to other RNNs such as LSTM, followed by a linear layer with weights $\mW_3$:
\begin{align}
\mathbf{H} = \mW_3 \left ( (\mathbf{C} + \vb_c) \odot \sigma(\mathbf{X}_2 + \vb_\sigma) \right ) + \vb_3
\end{align}
where $\sigma()$ is a gating activation function. We experimented with \texttt{sigmoid} activation and the \texttt{GeLU} activation~\citep{hendrycks2016bridging} used in BERT, and found the latter to achieve lower training loss. Finally, analogous to Eq.~\ref{eq:orig-transformer}, we set $\mX_{k+1} := F_{\textrm{Add+Norm}}(\mH, \bar{\mX}^k)$.

\paragraph{Speeding up recurrence}
In our experiments, we implement the recurrence step~(2) using the \texttt{scan()} function in Tensorflow.
Our model using this simple implementation runs 40\% slower than the standard Transformer, but is already much faster than other heavier RNNs such as LSTM.
For example, a Transformer model combined with LSTM can run multiple times slower~\cite{huang2020transblstm}.
\begin{table*}[!th]
\small
\centering
\begin{tabular}{lccccccccccc}
\toprule
  & BoolQ & CoLA & MNLI & MRPC & MultiRC & QNLI & QQP & RTE & SST2 & STSB & Avg \\
 \midrule
 \multicolumn{12}{c}{Base model (12 layers, $d=768$)} \\
 \hline
 BERT-orig & 73.3\% & \bf 82.0\% & 84.8\% & 88.5\% & 69.5\% & 91.0\% & 87.7\% & 64.0\% & 93.7\% & 84.2\% & 81.9\% \\
 BERT-rab & 70.1\% & 73.7\% & 85.4\% & 89.8\% & 70.3\% & 92.1\% & 87.5\% & 66.4\% & 91.1\% & 83.8\% & 81.0\% \\
 Ours & \bf 77.9\% & 80.8\% & \bf 85.9\% & \bf 90.4\% & \bf 74.2\% & \bf 92.5\% & \bf 88.2\% & \bf 70.8\% & \bf 93.8\% & \bf 85.7\% & \bf 84.0\% \\
 \midrule
 \multicolumn{12}{c}{Large model (24 layers, $d=1024$)} \\
 \hline
 BERT-orig & 84.5\% & 81.4\% & \bf 89.0\% & \bf 93.5\% & 79.6\% & 94.2\% & 88.6\% & 84.0\% & 95.3\% & \bf 87.9\% & 87.8\% \\
 BERT-rab & 84.5\% & 75.0\% & 88.9\% & 92.2\% & 80.9\% & 94.2\% & 88.4\% & 78.8\% & 93.1\% & 87.2\% & 86.3\% \\
 Ours & \bf 86.1\% &\bf 84.8\% & 88.9\% & 92.9\% & \bf 81.2\% & \bf 94.3\% & \bf 88.7\% & \bf 85.0\% & 95.3\% & 87.2\% & \bf 88.4\% \\
 \midrule
\multicolumn{12}{c}{Previously reported results (Large model)} \\
\hline
RoBERTa$^\dagger$ & - & 66.3\% & 89.0\% & 90.2\% & - & 93.9\% & 91.9\% & 84.5\% & 95.3\% & 91.6\% & - \\
BERT$^\dagger$ & - & 60.6\% & 86.6\% & 88.0\% & - & 92.3\% & 91.3\% & 70.4\% & 93.2\% & 90.0\% & - \\
BERT$^\ddag$ & - & 61.2\% & 86.6\% & 79.5\% & - & 93.1\% & 88.4\% & 68.9\% & 94.7\% & 89.6\% & - \\
 \bottomrule
\end{tabular}
\caption{Averaged development set results of all models. We perform 3 independent fine-tuning runs for each model and dataset. 
For comprehensive study, we also include previously reported results of large BERT models, although training details may differ in this and previous work.
Our baseline models are strong compared to previously reported results. $\dag$ indicate results from \citet{liu2019roberta} and $\ddag$ are results from \citet{alex2022mask} using an efficient training recipe and 40\% masking rate.\label{tab:overall}}
\end{table*}
\begin{figure*}[!th]
\centering
\includegraphics[width=6.35in]{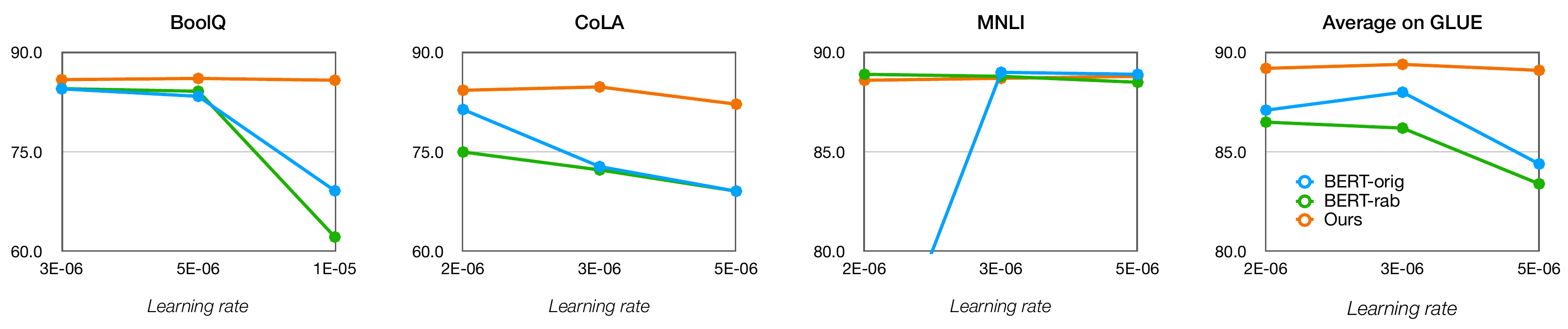}
\caption{Stability of fine-tuning results given different learning rates. Results are averaged across 3 independent runs for each setting. Our model is more robust to the range of learning rates tested.\label{fig:stability}}
\end{figure*}

We further improve the speed by increasing the step size for the RNN.
Specifically, $\mathbf{c}[i]$ is calculated using $\mathbf{c}[i-k]$ and $\mathbf{x}_1[i]$ with a step size $k > 1$.
Each recurrent step can process $k$ consecutive tokens at a time and only $\lceil l/k\rceil$ steps are needed.
In our experiments, we interleave the step size $k \in \{1, 2, 4\}$ across recurrent layers and found this to perform on par with using a fixed step size of 1.
Our model with variable step sizes has a marginal 20\% - 30\% slow-down compared to the standard Transformer model when training on TPUs.

Note that SwishRNN can be made significantly faster using optimized implementation such as CUDA kernel fusion adopted in QRNN and SRU~\cite{bradbury2016quasi,lei2018sru}.
We leave this for future work as custom kernel fusion is not readily available for TPUs.

\section{Experimental Setup}

\paragraph{Datasets}
Following BERT~\cite{devlin-etal-2019-bert}, we evaluate all models by pre-training them with the masked language model (MLM) objective and then fine-tuning them on a wide range of downstream tasks.
We use the Wikipedia and BookCorpus~\cite{ZhuKZSUTF15} for pre-training, and 10 datasets from the GLUE~\cite{wang-etal-2018-glue} and SuperGLUE benchmark~\cite{superglue}
including the BoolQ, CoLa, MNLI, MRPC, MultiRC, QNLI, QQP, RTE, SST2 and STS-B datasets.

\paragraph{Baselines}
We compare with two BERT variants. 
\textbf{BERT-orig} is the original BERT model using the multi-head attention described in~\citet{transformer} and learned absolute positional encoding.
The second variant \textbf{BERT-rab} adds the relative attention bias to each attention layer, following the T5 model~\cite{t5}.
Our model is the same as \textbf{BERT-rab} except we replace every FFN block with SwishRNN.
The inner hidden size $d'$ of SwishRNN blocks is decreased such that the total number of parameters are similar to the BERT baselines.
Following \citet{devlin-etal-2019-bert}, we experiment with two model sizes -- a base model setting consists of 12 Transformer layers and a large model setting using 24 layers.
The detailed model configurations are given in Appendix~\ref{sec:training_detail}.

\begin{table*}[!th]
\small
\centering
\begin{tabular}{lccccccccc}
\toprule
 Step size(s) of RNNs & CoLA & MNLI & MRPC & QNLI & QQP & RTE & SST2 & STSB & Time \\
 \midrule
 \multicolumn{10}{c}{Base model (12 layers, $d=768$)} \\
 \hline
 $\ 1$ & 79.4\% & 85.9\% & 90.8\% & 92.1\% & 88.1\% & 68.0\% & 92.7\% & 86.3\% & 1.4$\times$\\
 $\ 2$ & 75.5\% & 85.9\% & 88.6\% & 92.4\% & 88.1\% & 64.9\% & 92.1\% & 83.1\% & 1.2$\times$\\
 $\ \{1, 2, 4\}$ & 80.8\% & 85.9\% & 90.4\% & 92.5\% & 88.2\% & 70.8\% & 93.8\% & 85.7\% & 1.2$\times$\\
  \midrule
 \multicolumn{10}{c}{Large model (24 layers, $d=1024$)} \\
 \hline
 $\ 1$ & 84.7\% & 89.0\% & 92.5\% & 94.5\% & 88.5\% & 83.9\% & 95.6\% & 87.9\% & 1.4$\times$\\
 $\ \{1, 2, 4\}$ & 84.8\% & 88.9\% & 92.9\% & 94.3\% & 88.7\% & 85.0\% & 95.3\% & 87.2\% & 1.3$\times$\\
 \bottomrule
\end{tabular}
\caption{Fine-tuning results on the GLUE datasets using different step sizes for the recurrent module in our model. We report averaged results and the pre-training time in relative to that of BERT-rab model. Using variable step sizes trains faster and obtains results on par with using step size 1.\label{tab:stepsize}}
\end{table*}

\paragraph{Training}
Our pre-training recipe is similar to recent work~\cite{liu2019roberta,izsak-etal-2021-train,alex2022mask}.
Specifically, we do not use the next sentence prediction objective and simply replace 15\% input tokens with the special \texttt{[MASK]} token.
We also use a larger batch size and fewer training steps following recent work.
Specifically, we use a batch size of 1024 for base models and 4096 for large models.
The maximum number of pre-training steps is set to 300K.

To reduce the variance, we run 3 independent fine-tuning trials for every model and fine-tuning task, and report the averaged results.
We also tune the learning rate separately for each model and fine-tuning task.
The training details are provided in Appendix~\ref{sec:training_detail}.

\section{Results}
\paragraph{Overall results}
Table~\ref{tab:overall} presents the fine-tuning results on 10 datasets.
Our base model achieves a substantial improvement, outperforming the BERT-orig and BERT-rab baselines with an average of 2.1 absolute points.
The improvement is also consistent, as our base model is better on 9 out of the 10 datasets.

The improvement on large model setting is smaller.
Our model obtains an increase of 0.6 point and is better on 6 datasets.
We hypothesize that the increased modeling power due to recurrence can be saturating, as making the model much deeper and wider can already enhance the modeling capacity.
The gains are still apparent on more challenging datasets such as BoolQ where the input sequences are much longer.

\paragraph{Stability}
One interesting observation in our experiments is that combining recurrence and attention improves fine-tuning stability.
Figure~\ref{fig:stability} analyzes model stability by varying the learning rate.
We showcase the results on the first 3 datasets (namely BoolQ, CoLA and MNLI) as well as the averaged results on 8 datasets in GLUE.
For both BERT model variants, fine-tuning requires more careful tuning of the learning rate.
In comparison, our model performs much more consistently across the learning rates tested.

\paragraph{Step size of RNN}
Table~\ref{tab:stepsize} shows the effect of changing the step size of SwishRNN.
Using a step size of 1 is the slowest, since running smaller and more steps adds computational overhead.
On the other hand, using a fixed step size of 2 reduces the training cost but hurts the fine-tuning results especially on the CoLA, MRPC, RTE and STS-B datasets.
Our best model alternates the step size between 1, 2, and 4 across the recurrent layers, resulting in both faster training and stronger results.

\begin{figure}[!t]
\centering
\includegraphics[width=2.3in]{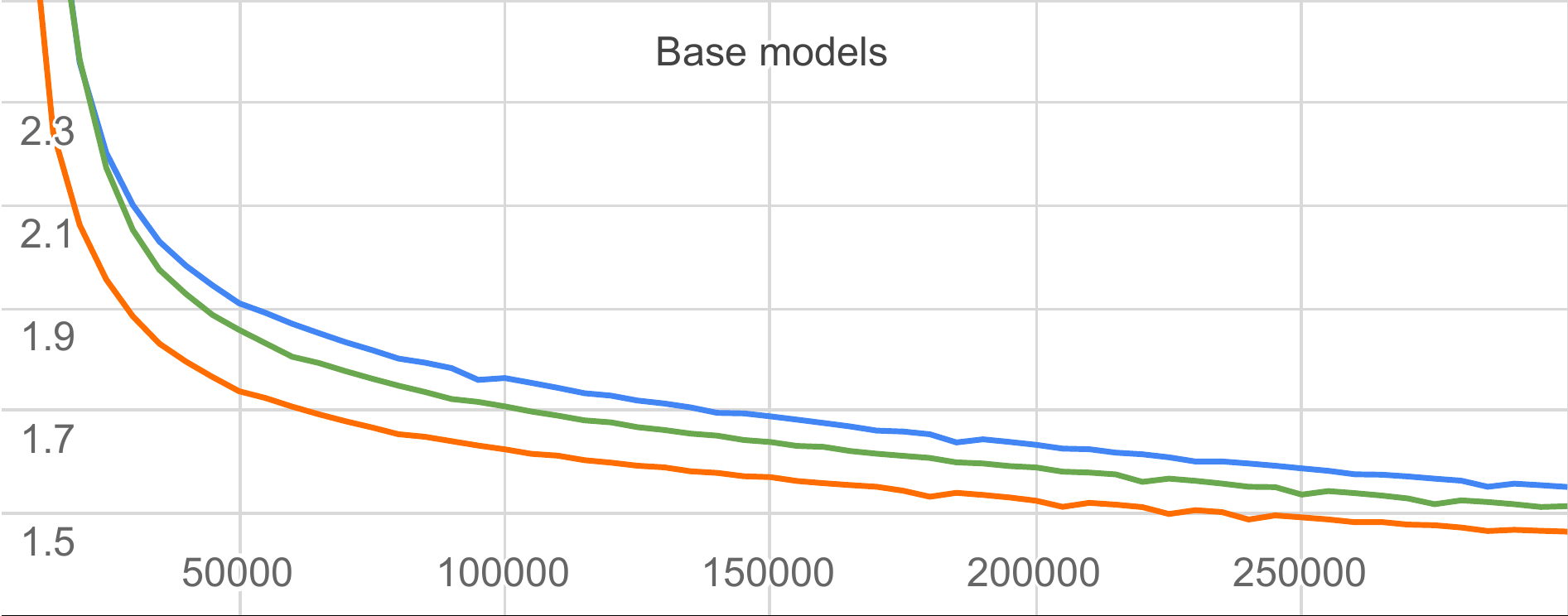} \\
\includegraphics[width=2.3in]{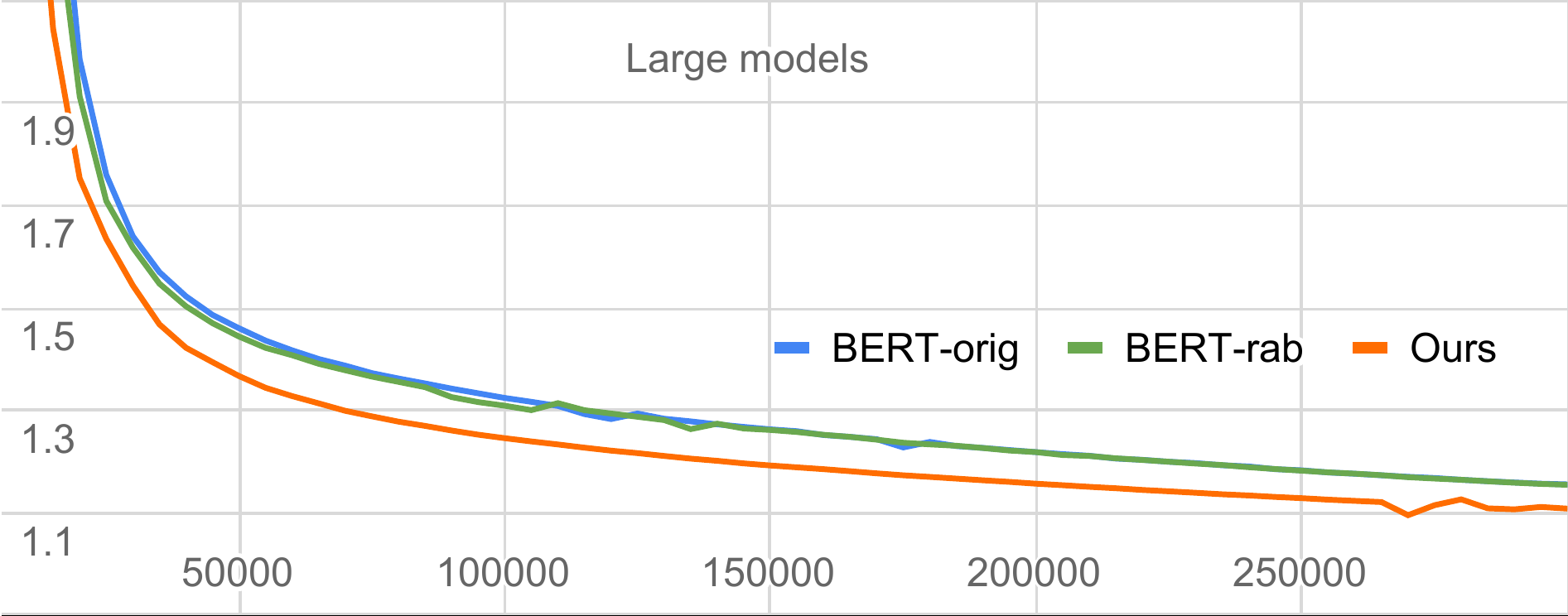}
\caption{MLM pre-training loss of BERT-orig, BERT-rab and our model architectures.\label{fig:loss}}
\end{figure}

\paragraph{Pre-training loss}
Figure~\ref{fig:loss} shows the training curves of all models during masked language model training.
Our model achieves better loss, indicating increased modeling capacity.

\section{Conclusion}

In this work, we proposed incorporating an extremely simple recurrent module, SwishRNN, that when incorporated into BERT achieves consistent improvements without requiring low-level performance optimizations. Future directions include extending our work to encoder-decoder pretraining~\cite{Song2019MASSMS} and exploring other domains such as protein modeling~\cite{Elnaggar2020ProtTransTC}.

\bibliography{custom}
\bibliographystyle{acl_natbib}

\appendix

\section{Transformer Architecture}

For completeness we review the $F_\textrm{att}$, $F_\textrm{ffn}$, and ${F}_{\textrm{Add+Norm}}$ blocks used in the Transformer architecture (Figure~\ref{fig:architecture}). We omit all bias terms for simplicity.
 
\paragraph{Attention block ($F_{\textrm{att}}$)} Multi-headed attention with $h$ heads first calculates query $\mQ_m$, key $\mK_m$, and value $\mV_m$ matrices for each head $m \in \{1,..h\}$ by applying linear transformations to the input:
\begin{align*}
\mQ_m & =\mX \mW^{Q}_m,\; \mK_m =\mX \mW^{K}_m,\; \mV_m =\mX \mW^{V}_m
\end{align*}
Each transformation matrix $\mW^{Q}_m, \mW^{K}_m, \mW^{V}_m$ is of dimension $d \times d_h$ where $d_h = d / h$. 
Attention vectors are then computed for each head, concatenated and multiplied by a linear transformation $\mW^O$ of dimension $d \times d$:
\begin{align*}
\mZ_m &= \softmax \left ( \frac{\mQ_m \mK^\top_m}{\sqrt{d_h}} \right ) \mV_m \\
F_{\textrm{att}}(\mX) &= \textrm{Concat}([\mZ_1,..., \mZ_h])\mW^O
\end{align*}

\paragraph{Feed forward block ($F_{\textrm{ffn}}$)} Following BERT, we use a \texttt{GeLU} nonlinearity~\citep{hendrycks2016bridging} i.e.
$F_{\textrm{ffn}}(\mX) =  \mW_{f2} (\texttt{GeLU} \left ( \mW_{f1} \mX) \right )$.

\paragraph{Residual connection and layer normalization ($F_{\textrm{Add+Norm}}$)}. This block applies layer normalization~\citep{ba2016layer} to the addition of the two inputs: $F_{\textrm{Add+Norm}}(\tilde{\mX}, \mX) = \textrm{LayerNorm}(\tilde{\mX} + \mX)$.

\section{Training details}
\label{sec:training_detail}

\paragraph{Pre-training}
The detailed hyper-parameter configuration for BERT training is shown in Table~\ref{tab:ptconfig}.
The training recipe is based on previous works such as RoBERTa~\cite{liu2019roberta} and the 24-hour BERT~\cite{izsak-etal-2021-train}.
Specifically, compared to the original BERT training recipe which uses 1M training steps and a batch size of 256, the new recipe increases the batch size.
The models are trained with much fewer steps and a larger learning rate as a result, which reduces the overall training time.
We train base models using 16 TPU v4 chips and large models using 256 chips.
For fine-tuning we use only 1 or 2 TPU v4 chips respectively.

\begin{table*}[h]
\centering
\begin{tabular}{lcc}
\toprule
 & Base model & Large model \\
\midrule
Number of layers & 12 & 24 \\
Hidden size  & 768 & 1024 \\
Inner hidden size -- FFN & 3072 & 4096 \\
Inner hidden size -- SwishRNN & 2048 & 2752 \\
Attention heads & 12 & 16 \\
Attention head size & 64 & 64 \\
Dropout & 0.1 & 0.1 \\
Attention dropout & 0.1 & 0.1 \\
Learning rate & 0.0003 & 0.0002 \\
Learning rate warmup steps & 20,000 & 20,000\\
Learning rate decay & Linear & Linear \\
Adam $\beta_1$ & 0.9 & 0.9 \\
Adam $\beta_2$ & 0.98 & 0.98 \\
Weight decay & 0.01 & 0.01 \\
Batch size & 1024 & 4096 \\
Sequence length & 512 & 512 \\
Training steps & 300,000 & 300,000 \\
\bottomrule
\end{tabular}
\caption{Hyper-parameters for pre-training the base models and large models in our experiments.\label{tab:ptconfig}}
\end{table*}

\paragraph{Fine-tuning}
We use a batch size of 32 for fine-tuning and evaluate the model performance every 1000 steps.
We use Adam optimizer without weight decay during fine-tuning.
We use a fixed learning rate tuned among $\{\text{1e-5}, \text{5e-6}, \text{3e-6}, \text{2e-6}\}$ and warm up the learning rate for 1000 steps.
The maximum number of training steps of each dataset is presented in Table~\ref{tab:ftconfig}.
We set the number proportionally to the size of the dataset and do not tune it in our experiments.

\begin{table*}[ht]
\centering
\begin{tabular}{cccccccccc}
\toprule
 BoolQ & CoLA & MNLI & MRPC & MultiRC & QNLI & QQP & RTE & SST2 & STSB \\
 \midrule
 50K & 50K & 200K & 20K & 50K & 100K & 150K & 20K & 80K & 30K \\
 \bottomrule
\end{tabular}
\caption{Maximum number of fine-tuning step used for each dataset in our experiments.\label{tab:ftconfig}}
\end{table*}

\end{document}